\documentclass{article}

\usepackage{arxiv}

\usepackage[utf8]{inputenc} 
\usepackage[T1]{fontenc}    
\usepackage{hyperref}       
\hypersetup{
    colorlinks=true,
    linkcolor=cyan,
    filecolor=magenta,      
    urlcolor=blue,
}
\usepackage{url}            
\usepackage{booktabs}       
\usepackage{amsfonts}       
\usepackage{nicefrac}       
\usepackage{microtype}      
\usepackage{lipsum}
\usepackage{graphicx}
\usepackage{amsmath}
\usepackage{multirow}
\usepackage{newtxtext,newtxmath}
\DeclareMathOperator*{\argmax}{Arg\,Max}

\title{Compressing deep neural networks on FPGAs to binary and ternary precision with \texttt{hls4ml}}

\author{
  Jennifer Ngadiuba, Vladimir Loncar~\thanks{Also at Institute of Physics Belgrade, Serbia.}, Maurizio Pierini, Sioni Summers \\
  European Organization for Nuclear Research (CERN) \\
  CH-1211 Geneva 23, Switzerland
  \And
  Giuseppe Di Guglielmo \\
  Columbia University \\
  New York, NY 10027, USA
  \And
  Javier Duarte \\
  University of California San Diego \\
  La Jolla, CA 92093, USA \\
  \And 
  Philip Harris, Dylan Rankin\\
  Massachusetts Institute of Technology\\
  Cambridge, MA 02139, USA 
  \And
  Sergo Jindariani, Mia Liu, Kevin Pedro, Nhan Tran \\
  Fermi National Accelerator Laboratory\\
  Batavia, IL 60510, USA\\
  \And
  Edward Kreinar\\
  HawkEye360\\
  Herndon, VA 20170, USA\\
  \And
  Sheila Sagear\\
  Boston University\\
  Boston, MA 02215, USA
  \And
  Zhenbin Wu\\
  University of Illinois at Chicago\\
  Chicago, IL 60607, USA
  \And
  Duc Hoang\\
  Rhodes College\\
  Memphis, TN 38112, USA   
}

\begin{document}
\begin{center}
\includegraphics[width=8cm]{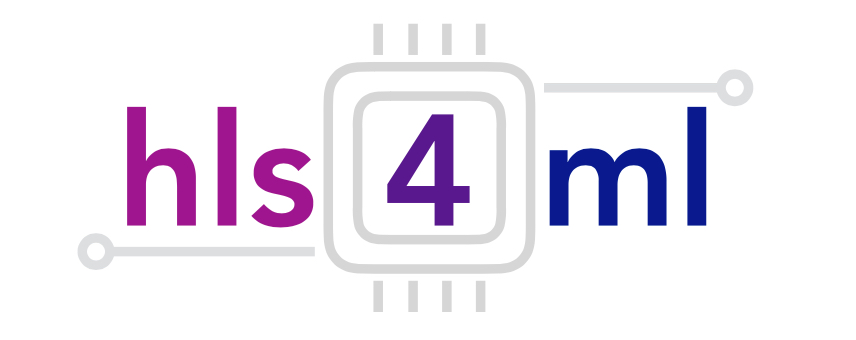}
\end{center}
\maketitle

\begin{abstract}
  We present the implementation of binary and ternary  neural networks in the {\tt hls4ml} library, designed to automatically convert deep neural network models to digital circuits with FPGA firmware.  Starting from benchmark models trained with floating point precision, we investigate different strategies to reduce the network's resource consumption by reducing the numerical precision of the network parameters to binary or ternary. We discuss the trade-off between model accuracy and resource consumption. In addition, we show how to balance between latency and accuracy by retaining full precision on a selected subset of network components. As an example, we consider two multiclass classification tasks: handwritten digit recognition with the MNIST data set and jet identification with simulated proton-proton collisions at the CERN Large Hadron Collider. The binary and ternary implementation has similar performance to the higher precision implementation while using drastically fewer FPGA resources.
\end{abstract}

\newpage

\section{Introduction}

Field-programmable gate arrays (FPGAs) are an efficient and flexible processing
solution to perform low latency and high bandwidth inference of deep neural networks
(DNNs). Their design is extremely functional to parallelize the mathematical operations typical of DNN inference
tasks, namely matrix multiplication and activation function application. FPGAs can be reprogrammed, which offers advantages in terms of flexibility with respect to application-specific integrated circuits (ASICs). At the same time, they share some of the advantages offered by ASICs, such as low power consumption and speed.

Typically, FPGAs are used to emulate generic digital circuits as a preliminary step toward the design of custom ASICs or as an
alternative to them. For instance, hundreds of FPGAs are used as custom electronic logic to process in real time the proton-proton
collisions at the CERN Large Hadron Collider (LHC). With beams
colliding every 25~ns and thanks to a built-in buffering system, a typical LHC experiment has ${\cal O}(1)~\mu$s to decide whether to keep or discard a given event. This real-time decision-taking system, referred to as the level-1 (L1) trigger, consists of a set of digital circuits implementing physics-motivated rule-based selection algorithms. Currently, these algorithms are deployed on FPGAs, mounted on custom electronics boards.

The severe L1 latency constraint prevents the LHC experimental collaborations
from deploying complex rule-based algorithms on the L1 FPGA boards. Machine Learning (ML) solutions, and in particular DNNs, are currently being investigated as fast-to-execute and parallelisable approximations of rule-based algorithms.
For instance, the CMS collaboration has deployed boosted decision trees (BDTs) in the L1 trigger electronic logic~\cite{cms_l1t_bdt}. Following this approach, one could train a DNN to process a given input (e.g., energy deposits in a calorimeter) and return the output of an event reconstruction algorithm (e.g., to regress the energy of the incoming particle that caused these energy deposits or to identify its nature). Because the complexity of LHC collision events is going to increase after the upcoming high-luminosity upgrade, we expect this approach to become more prevalent.

In order to  facilitate the deployment of DNNs in the L1 trigger systems of high energy physics (HEP) experiments, we developed a software library, {\tt hls4ml}, to convert a DNN model into FPGA firmware through an automatic workflow~\cite{Duarte:2018ite}. In HEP, the deployment of deep learning (DL) models on FPGAs has been discussed in the context of the online data-selection system of the LHC experiments. Alternative solutions based on VHDL~\cite{Schlag:2670301} have been explored. Similar studies and comparable results have been shown in Ref.~\cite{s19132981}. 

The {\tt hls4ml} design is characterized by two aspects: (i) a reliance on high-level synthesis (HLS) backends, in order to fully automate the workflow from a trained model to FPGA firmware; (ii) a target of fully-on-chip logic, which enables the latency to be within typical values of ${\cal O}(1)~\mu$s. Our ultimate goal is to support the most popular DNN model ingredients (layers, activation functions, etc.) and an interface to the most popular DL training libraries, directly (e.g., for {\tt TensorFlow}~\cite{TF}, {\tt Keras}~\cite{keras}, and {\tt PyTorch}~\cite{pytorch}) or through the {\tt ONNX}~\cite{onnx}
interface. The library is under development and many of these ingredients are already supported. While {\tt hls4ml} was initially conceived for LHC applications,
its potential use cases go well beyond HEP. In general, {\tt hls4ml} provides a user-friendly interface to deploy custom DNN models on FPGAs, used as co-processing accelerators or as digital circuits in resource-constrained, low-latency computing environments.

In addition, the {\tt hls4ml} library supports the deployment of BDTs on FPGAs~\cite{Summers:2020xiy}. A BDT trained on high-level features can often reach similar performances than small fully-connected neural networks. On the other hand, neural networks offer the possibility to directly process the raw data, saving time and resources that would be otherwise spent to compute the input features. Depending on the use case, a developer would decide which workflow better fits her needs.

The main challenge in deploying a DNN model on an FPGA is the limited
computational resources. Typically, one would reuse resources
for the inference operations across multiple clock cycles,
at the price of a larger latency. The \emph{reuse factor} quantifies how many times a resource is reused and is equal to the initiation interval (II) for that operation. A complementary approach consists of compressing the model, e.g., by reducing the number of operations needed in the inference step (pruning) or their cost (e.g., quantizing the network using a reduced numerical representation). Comprehensive reviews of these techniques can be found in Ref.~\cite{DBLP:journals/spm/ChengWZZ18,compress}. In a previous publication~\cite{Duarte:2018ite}, we showed that pruning~\cite{DBLP:journals/corr/HanMD15,DBLP:journals/corr/HanPTD15} and quantization~\cite{DBLP:journals/corr/HanMD15,DBLP:journals/corr/LinTA15} allow one to execute simple fully-connected DNN models with state-of-the-art performance on a specific LHC problem within a latency of ${\cal O}(100)$~ns, while using only a fraction of the FPGA resources. In this paper, we investigate how a similar result can be obtained with binary and ternary networks~\cite{courbariaux,binary_first,ternary_first}, following closely the studies presented in Refs.~\cite{courbariaux,xilinx_finn,bertmoons}. Network parameters in binary (ternary) networks assume values $+1$ or $-1$ ($+1$, $0$, or $-1$). They can be represented with one bit (two bits), resulting in a much smaller resource consumption.

In this study, we consider two benchmark problems: MNIST digit classification, which allows a direct comparison with previous literature~\cite{xilinx_finn}; the jet tagging problem used as benchmark in our previous study~\cite{Duarte:2018ite} as well as by other groups~\cite{s19132981}. The jet tagging problem is particularly relevant for applications at the LHC. Traditional algorithms for jet tagging are too complex to run within L1 latency constraint. Developing resource-friendly ultrafast solutions for jet tagging would drastically increase the L1 selection quality for all-jet collision events. One should keep in mind that our LHC jet data set represents a simplification of more complex realistic conditions. It does not take into account the time and resources one would spend to compute the input features. In the future, the extension of the hls4ml library to more complex architectures will allow to consider more realistic use cases, with raw data being directly processes by compressed models.

This paper is structured as follows:  Section~\ref{sec:data} introduces the benchmark problems and data sets. The implementation of binary and ternary networks in {\tt hls4ml} is described in Section~\ref{sec:FPGAopt}. Section~\ref{sec:architectures} describes the different model architectures considered in this study, while their application to the two benchmark classification problems is discussed in Section~\ref{sec:exp}. The summary and outlook are given in Section~\ref{sec:conclusions}.

\section{Benchmark models and data sets}
\label{sec:data}

We consider two benchmark classification tasks: a digit recognition task with the MNIST data set~\cite{MNISTdata} and the LHC jet tagging task discussed in Ref.~\cite{Duarte:2018ite}. 

\begin{figure}[th!]
    \centering
    \includegraphics[width=\textwidth]{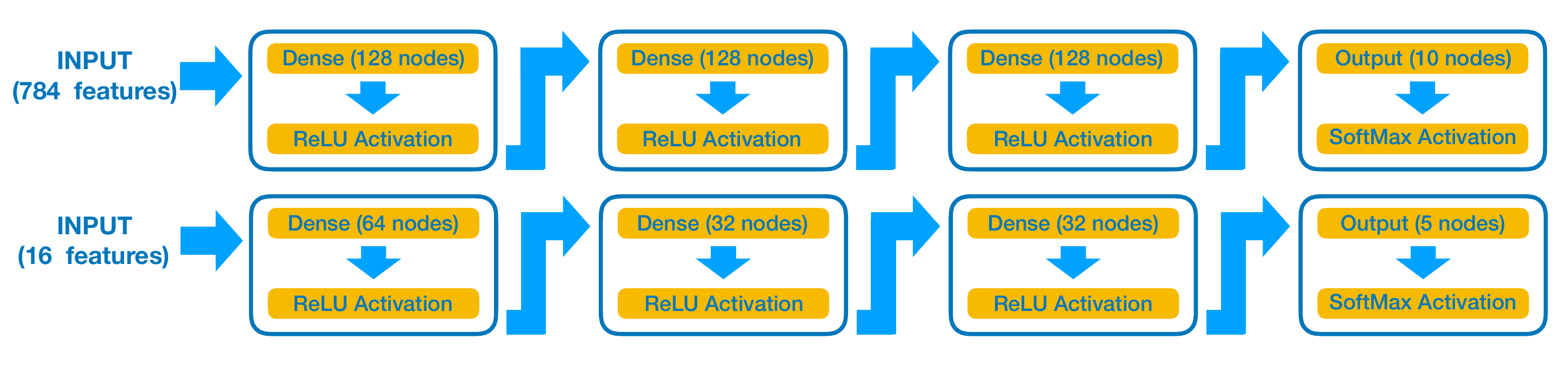}
    \caption{Network architecture for the baseline MNIST (top) and LHC jet (bottom) classifiers used as benchmark models in this study.\label{fig:input_arch}}
\end{figure}

\begin{figure}[th!]
    \centering
    \includegraphics[width=0.45\textwidth]{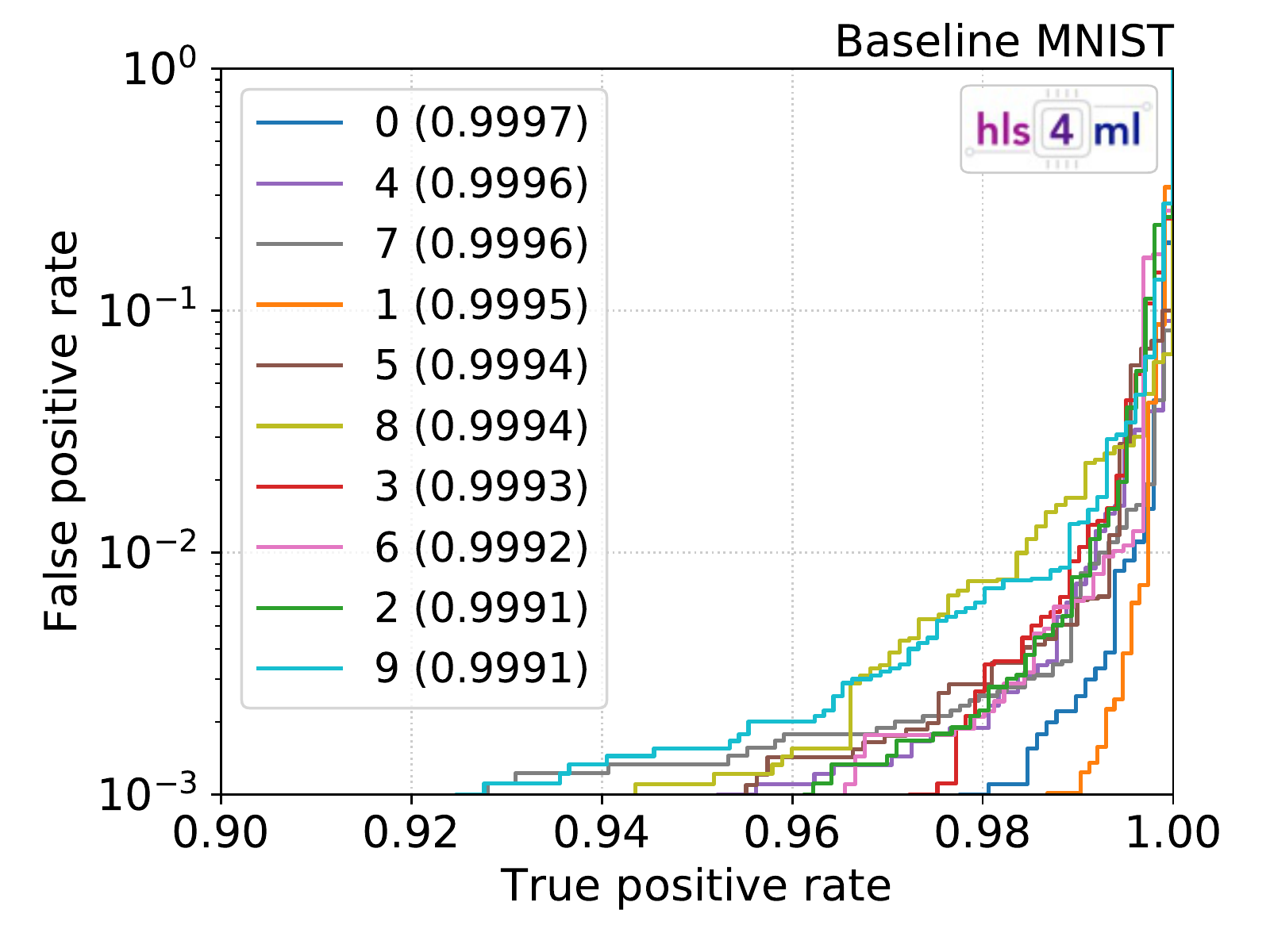}
    \includegraphics[width=0.45\textwidth]{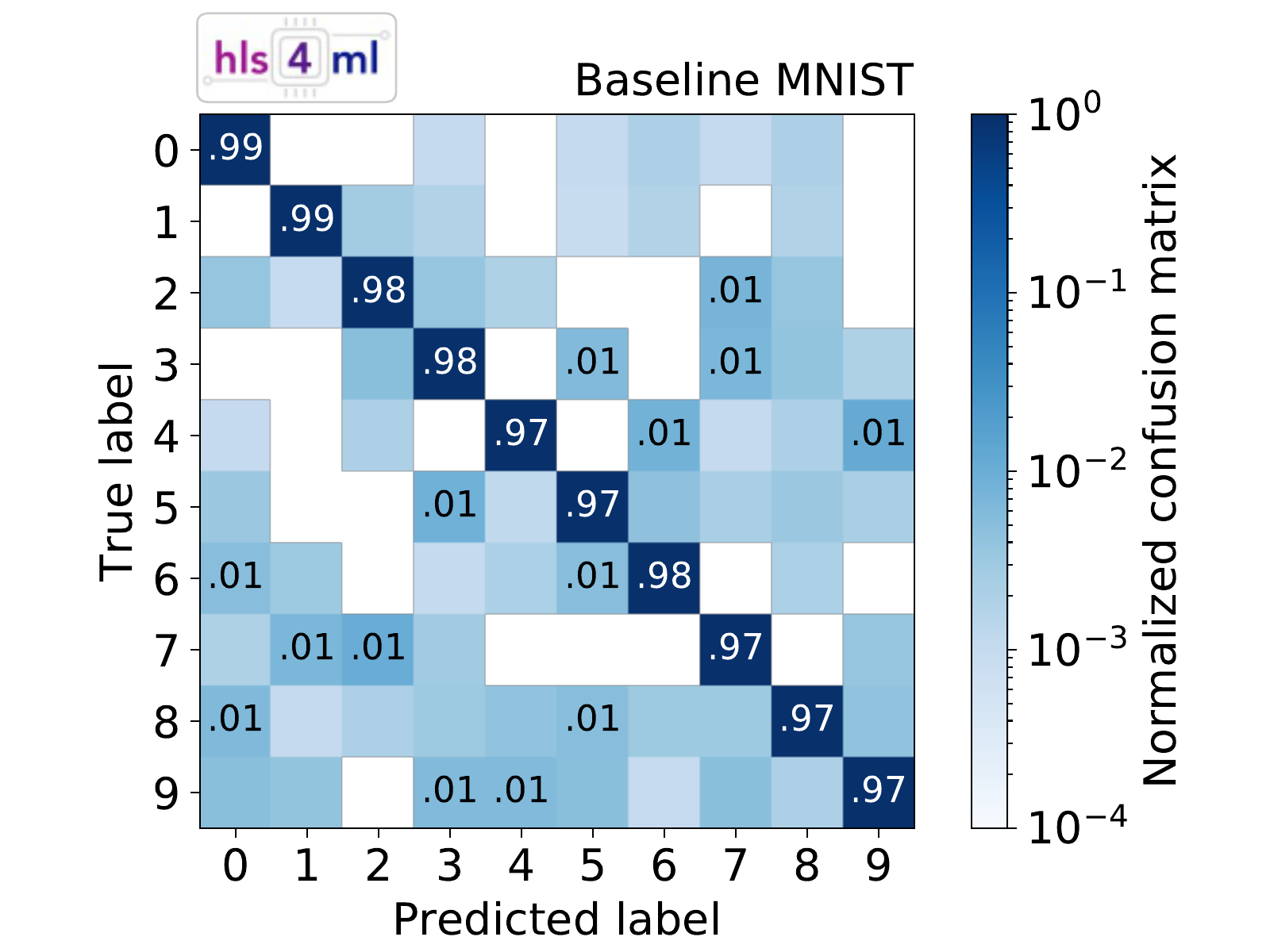} \\
    \includegraphics[width=0.45\textwidth]{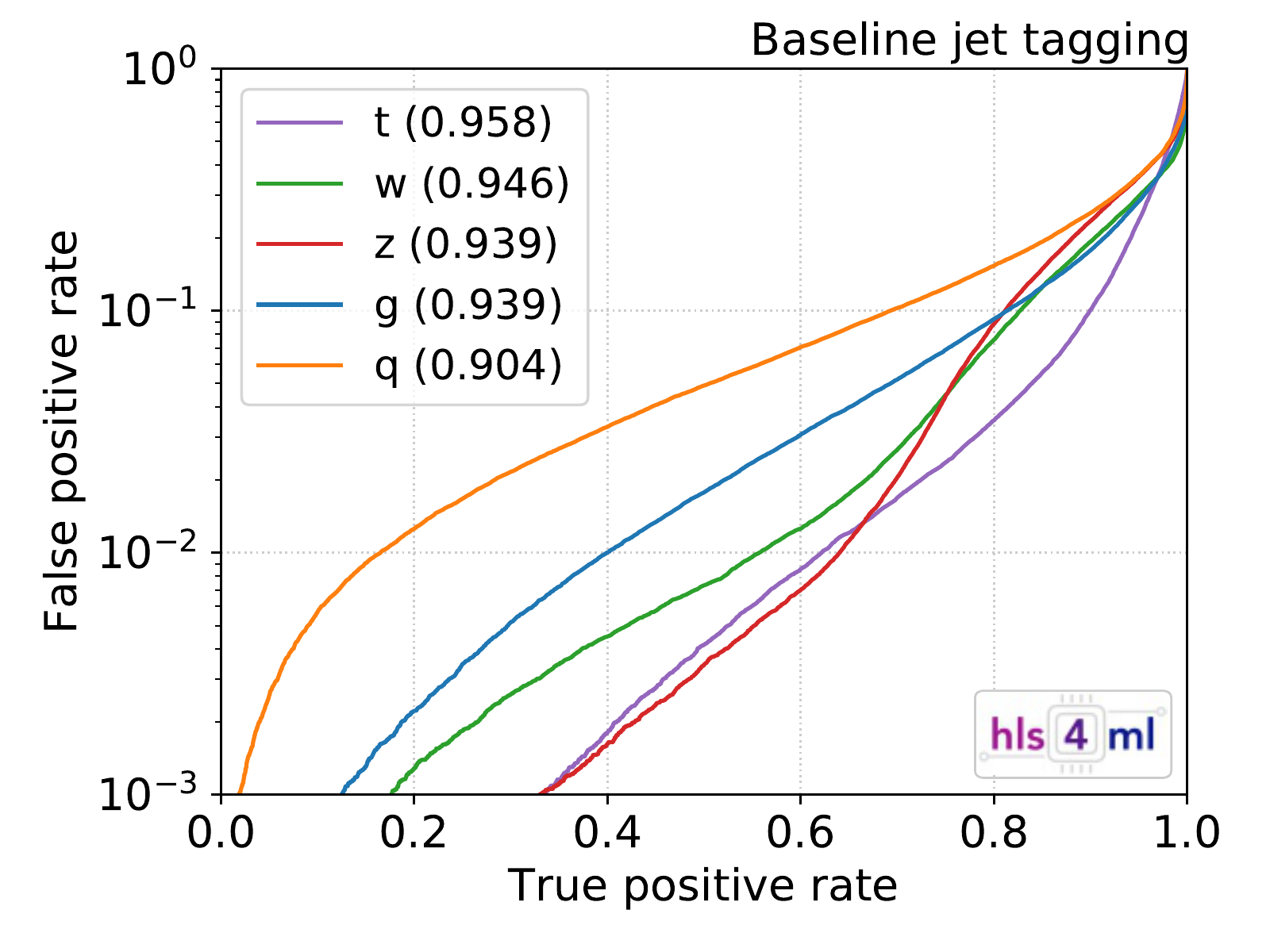}
    \includegraphics[width=0.45\textwidth]{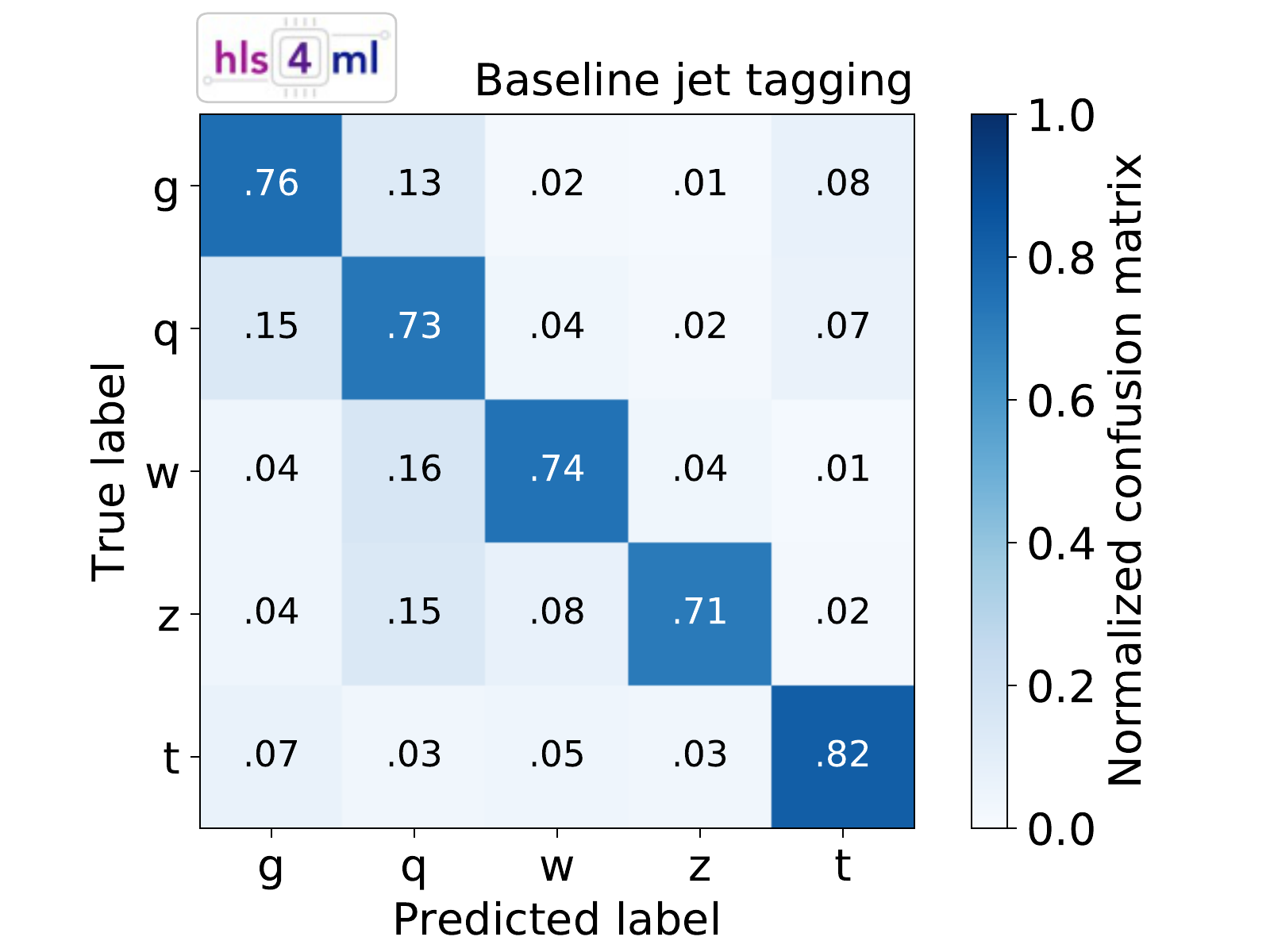}
    \caption{Classification performance evaluated on the testing sample of the baseline MNIST (top) and LHC jet (bottom) classifiers used as benchmark models in this study: ROC curves (left) and normalized confusion matrices (right). On the left, numbers in parentheses correspond to the AUC of each class. On the right, the text is omitted for bins corresponding to a false positive rate below 1\%.\label{fig:float_perf}}
\end{figure}

The MNIST data set consists of images of hand-written digits. Each image is represented as a $28\times 28$ pixel array, storing the gray-scale content of each pixel in the original image. For our purpose, we flatten the 2D array to a 1D array, concatenating each row of the image to the right to the previous one. The derived 1D array is passed as input to a multilayer perceptron (MLP)~\cite{MLP} with an input (output) layer of 784 (10) nodes and three hidden layers with 128 nodes each. Rectified linear unit (ReLU) activation functions~\cite{Hahnloser:2000wt} are used for the hidden layer nodes, while a softmax activation function is used for the output layer. The MNIST data set comes divided into training-and-validation samples (with 60,000 images) and a testing samples (with 10,000 images). We use 75\% of the training-and-validation data set for training, and the remaining 25\% for validation.

The other benchmark task consists of classifying jets from a set of 16 physics-motivated high-level features, as described in Refs.~\cite{Duarte:2018ite,Coleman:2017fiq}. The input data set consists of simulated jets with an energy of order 1 TeV, originating from light quarks ($q$), gluons ($g$), $W$ bosons, $Z$ bosons, or top quarks ($t$) produced in proton-proton collisions at a center-of-mass energy of 13 TeV. Jets are clustered using the anti-$k_\mathrm{T}$ algorithm~\cite{Cacciari:2008gp}, with distance parameter $R=0.8$. For each jet, the 16 high level features are computed and given as input to a multiclass MLP classifier. The data set is available in the Zenodo repository~\cite{data}. More details on the data set can be found in Refs.~\cite{Coleman:2017fiq,Duarte:2018ite,Moreno:2019bmu}. The data set consists of approximately 1 million examples and is split in three parts: 20\% for test, 60\% for training, and 20\% for validation.
The network receives as input the 16 high-level features and processes them through a MLP with three hidden layers of 64, 32, and 32 nodes with ReLU activation functions. The output layer consists of five nodes with softmax activation. The five output values correspond to the probability that a given jet belongs to one of the five jet classes. 

The architectures of the baseline MNIST and LHC jet classifiers are illustrated in Fig.~\ref{fig:input_arch}. Both are implemented and trained with {\tt Keras} in floating point precision (FPP). Their performance is shown in Fig.~\ref{fig:float_perf} in terms of receiver operating characteristic (ROC) curves and normalized confusion matrices. The area under the curve (AUC) of each ROC curve is quoted in the figure, as well as in Table~\ref{tab:performance_benchmark}, where the corresponding accuracy values are also given. Following convention, we define the model accuracy as the fraction of correctly labeled examples, also referred to as true positives (TP)

\begin{equation}
\frac{\sum_{i=1}^{C} \mathrm{TP}_{i}}{N} ,
\end{equation}

where the sum runs over the number of classes $C$ and $N$ is the total number of examples. The accuracy per class is calculated taking into account also the true negatives (TN), i.e. the examples not belonging to that class and that have been predicted in one of the other classes

\begin{equation}
\frac{\sum_{i=1}^{C} \mathrm{TP}_{i} + \mathrm{TN}_{i}}{N} .
\end{equation}

In practice, the computation of the model or per-class accuracy is done applying an $\argmax$ function to the array of scores returned by the network and comparing it to the corresponding target array. The total accuracy of the MNIST and LHC jet classifiers, computed across all categories, are found to be 98\% and 75\%, respectively.

\begin{table}[ht!]
  \caption{Classification performance evaluated on the testing sample of the baseline MNIST and LHC jet classifiers used as benchmark models in this study: AUC and per-class accuracy.\label{tab:performance_benchmark}}
  \centering
\begin{tabular}{c|cc|c|cc}
\hline
\multirow{2}{*}{Class} & \multicolumn{2}{|c|}{MNIST} & \multirow{2}{*}{Class} & \multicolumn{2}{|c}{Jet tagging} \\
&  AUC & Accuracy [\%] &  & AUC & Accuracy [\%] \\
\hline
0 & 0.9997 & 99.7 & \multirow{2}{*}{$g$} & \multirow{2}{*}{0.939} & \multirow{2}{*}{89}\\
1 & 0.9995 & 99.8 & & & \\
2 & 0.9991 & 99.6 & \multirow{2}{*}{$q$} & \multirow{2}{*}{0.904} & \multirow{2}{*}{85}\\
3 & 0.9993 & 99.6 & & & \\
4 & 0.9996 & 99.6 & \multirow{2}{*}{$W$}   & \multirow{2}{*}{0.946} & \multirow{2}{*}{91}\\
5 & 0.9994 & 99.6 & & & \\
6 & 0.9992 & 99.6 & \multirow{2}{*}{$Z$}   & \multirow{2}{*}{0.939} & \multirow{2}{*}{92}\\
7 & 0.9996 & 99.6 & & & \\
8 & 0.9994 & 99.4 & \multirow{2}{*}{$t$}   & \multirow{2}{*}{0.958} & \multirow{2}{*}{93}\\
9 & 0.9991 & 99.5 & & & \\
\hline
\end{tabular}
\end{table} 

These baseline architectures were chosen in order to provide a reasonable performance while keeping the resource utilization within a manageable level. The state-of-the-art performance on MNIST reaches higher accuracy than the models considered here. However, these models are extremely lightweight in terms of their small number of parameters, and low precision. They are therefore optimized for their small footprint of resources and latency in the FPGA inference. Similarly, any jet classifier algorithm with accuracy $\sim 60-70$\%, like the one we consider, would be of great benefit for LHC experiments: since the majority of jets produced at the LHC comes from quarks and gluons, our baseline model would allow one to select $> 80\%$ of $W$, $Z$, and $t$ jets while reducing the required bandwidth by a factor $\sim 10$, saving resources that could be used to extend the physics program of the experiment in other directions.

We consider these models as examples, which are not intended to represent the best reachable performance for a given use case. No architecture optimization was attempted, since the focus of this study is on their implementation on hardware and relative performance drop rather than on absolute performance.

\section{Implementing binary and ternary networks in \texttt{hls4ml}}
\label{sec:FPGAopt}

\begin{figure}[t!]
\centering
\includegraphics[width=0.4\textwidth]{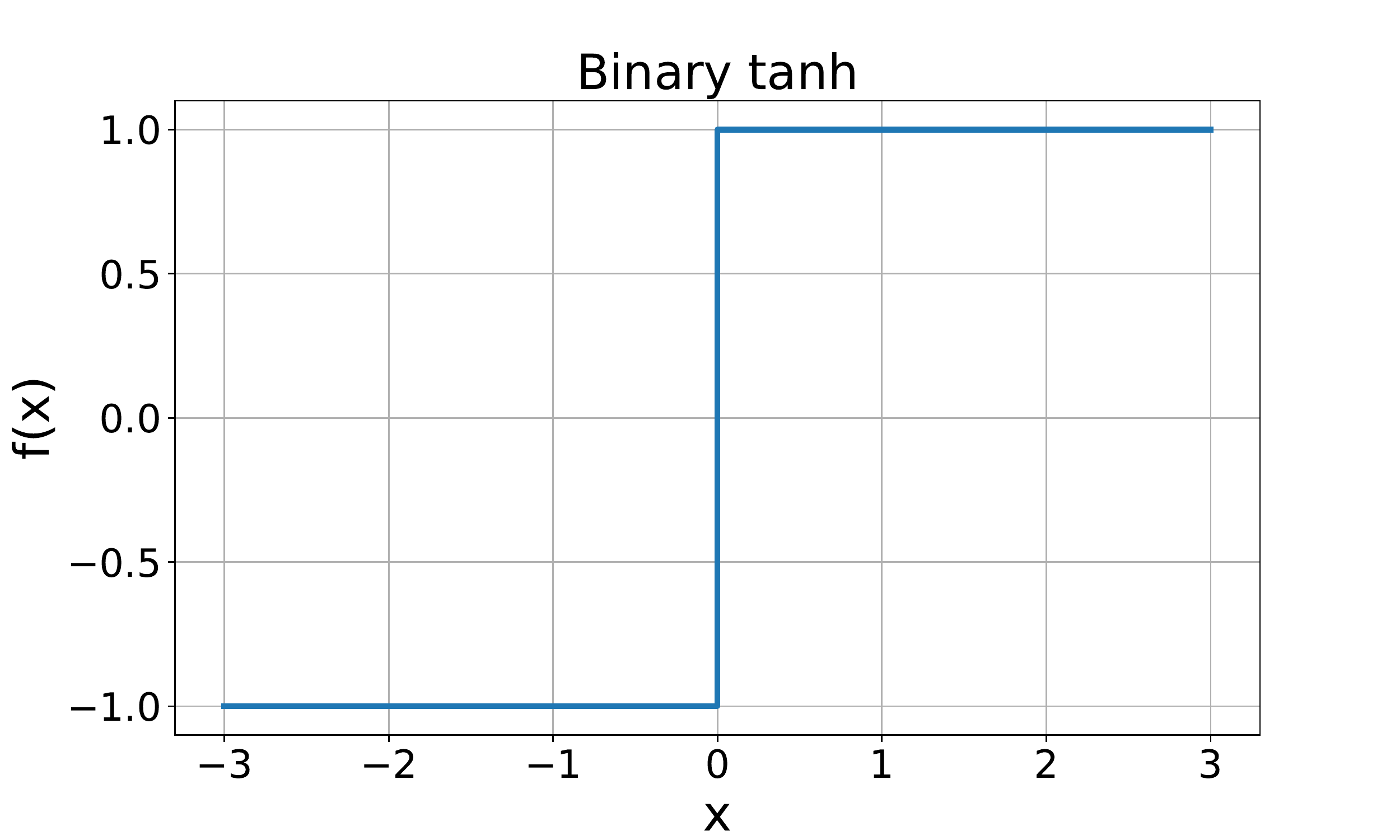}
\includegraphics[width=0.4\textwidth]{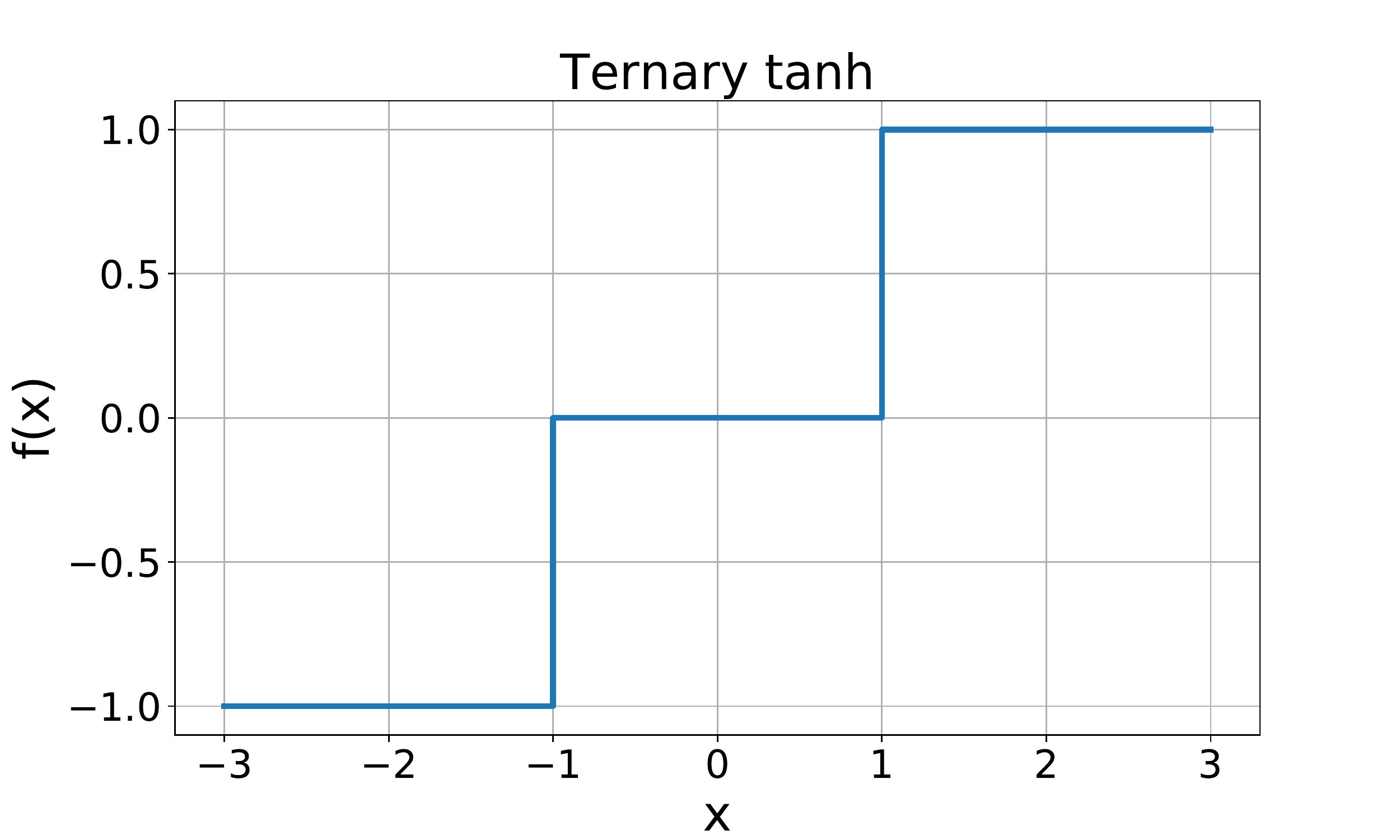} \\
\includegraphics[width=0.4\textwidth]{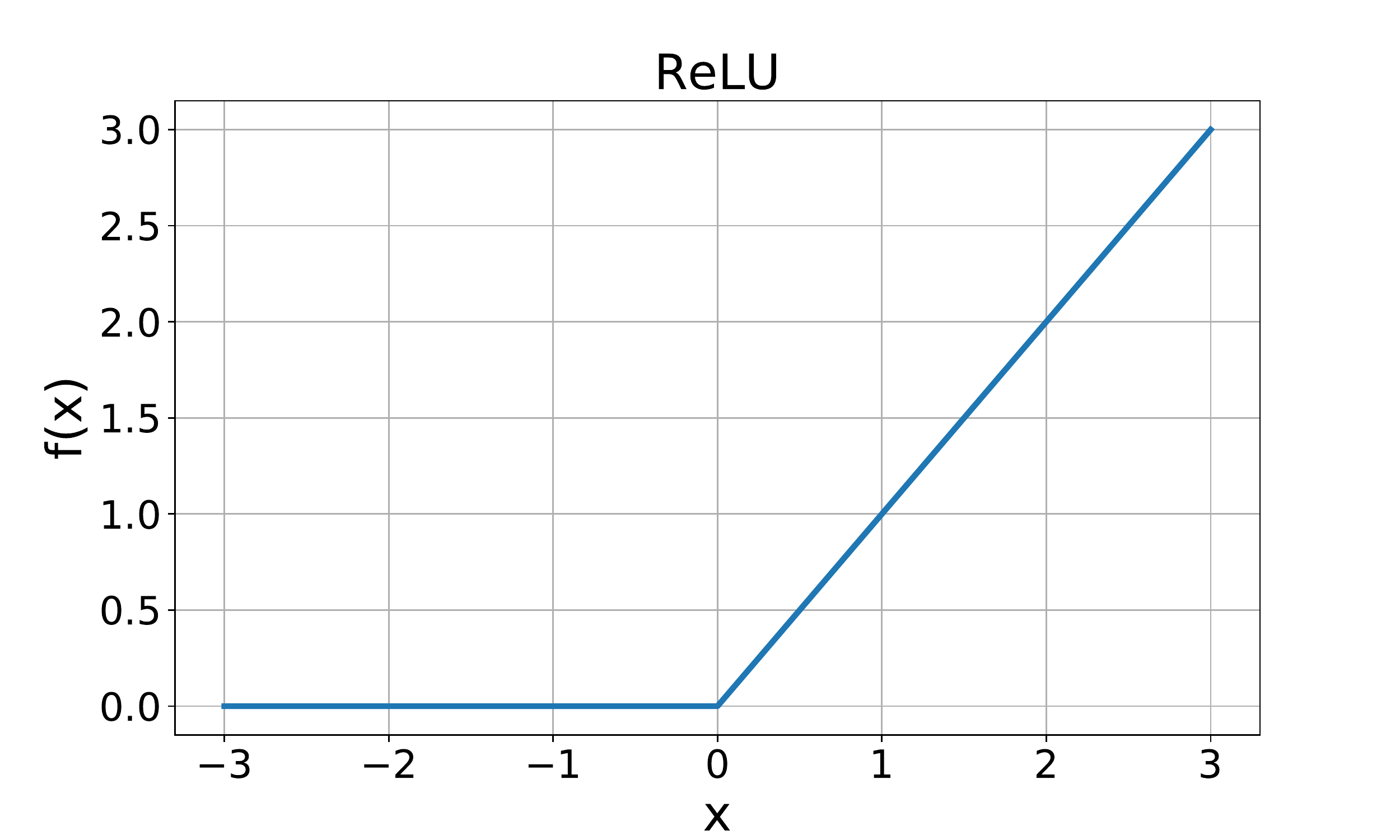}
\includegraphics[width=0.4\textwidth]{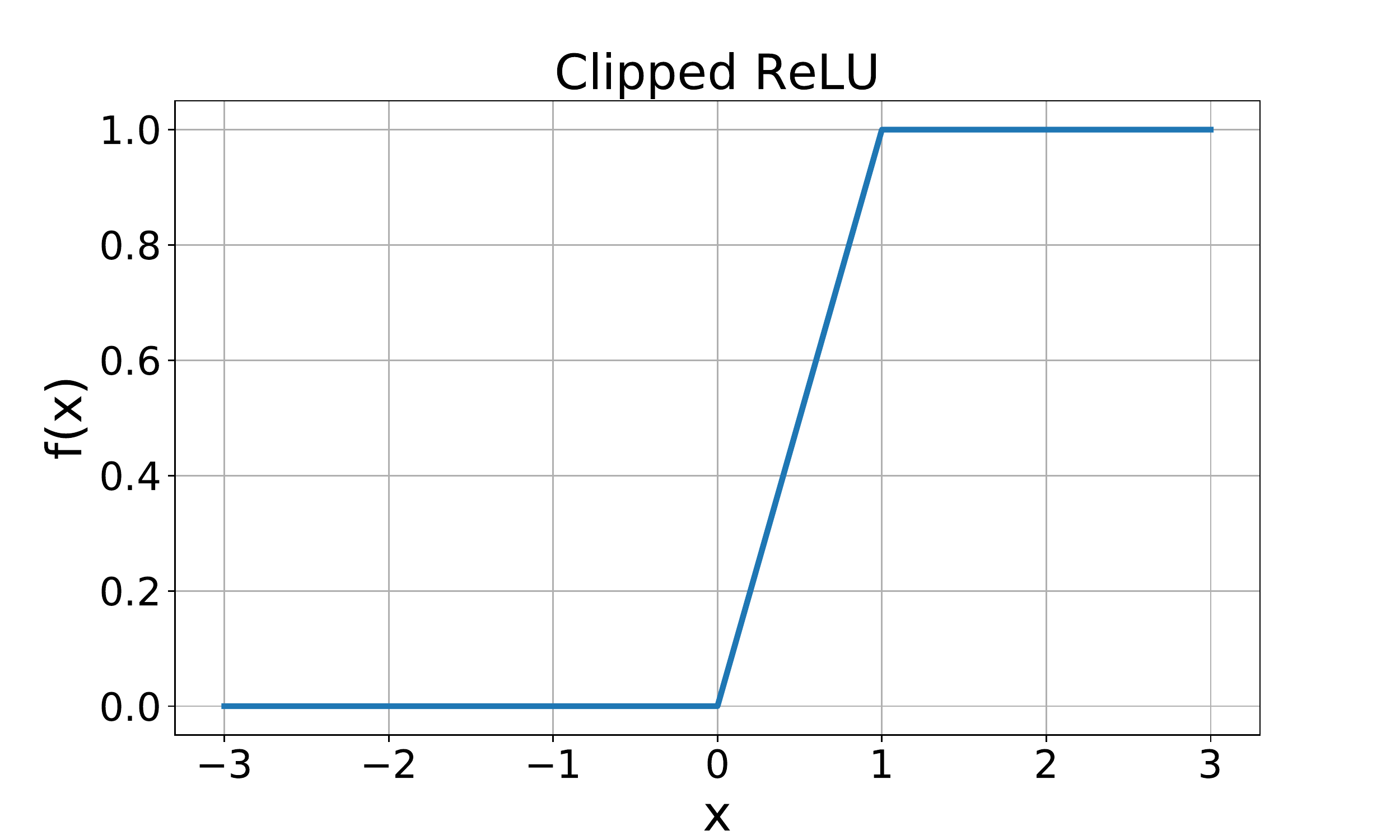} \\
\caption{Activation functions used to define the models described in
  Section~\ref{sec:architectures}: binary $\tanh$ (top-left), ternary $\tanh$
  (top-right), ReLU (bottom-left) and clipped ReLU
  (bottom-right).\label{fig:activation_binary_ternary}}
\end{figure}

Binary and ternary networks are extreme examples of quantized neural networks~\cite{Duarte:2018ite}. A network is quantized when its parameters (operations) are represented (performed) with reduced numerical precision. This precision could be the same across the full network or specific to each component (e.g., for different layers). Quantization reduces the computing resources of model inference and its level can be tuned to yield little or no loss in model performance. 
In the case of binary (ternary) networks, each weight assumes a value of $+1$ or $-1$ ($+1$, $0$, or $-1$). Two- and three-valued activation functions are used after each layer, acting as discrete versions of the $\tanh$ function. As alternatives, we also investigate a standard ReLU function as well as its clipped version~\cite{Cai2017DeepLW}, defined as $\min(\mathrm{ReLU}(x), y_\mathrm{max})$, with $y_\mathrm{max}$ being a positive hyperparameter. In our study, we fix $y_\mathrm{max}=1$. The four functions are shown in Fig.~\ref{fig:activation_binary_ternary}. 

In order to convert the models described in Sections~\ref{sec:data}, we rely on the MLP-related functionalities offered by the \texttt{hls4ml} library, discussed at length in Ref.~\cite{Duarte:2018ite}. In addition to that, we exploit a set of custom implementations~\cite{xilinx_finn}, specific to binary and ternary networks, that allow one to speed up the execution of the building-block architecture shown in Fig.~\ref{fig:architecture}. The implementation of these solutions is integrated in recent versions of the {\tt hls4ml} library, starting with the {\tt v0.1.6} tag of the GitHub repository~\cite{hls4ml_github}. With respect to the work presented in Ref.~\cite{Duarte:2018ite}, this version provides a special support for large dense layers containing hundreds of nodes as in the models we consider in this study. This functionality will be described in more detail in a future publication.

Binary networks use 1-bit representations for both weights and activations. In this case, the product between two quantities can be optimized as an extremely lightweight operation. By encoding an arithmetical value of `$-1$' as `$0$', the product can be expressed as an XNOR operation. As described in Table~\ref{tab:xnor_product}, an XNOR filter returns $0$ when the two input values are different and $1$ otherwise.
For models using ternary weights or greater than 1 bit for activations, the much larger FPGA logic is always used rather than digital signal processing (arithmetic) blocks (DSPs), whose number is typically limited.

\begin{table}[h]
\caption{Left: All possible products between $A$ and $B$ with values constrained to $\pm 1$. Right: The corresponding truth-table when the quantities $A$ and $B$ are each encoded with 1 bit, and the XNOR operation is used for the product.} 
\label{tab:xnor_product}
\centering
\begin{tabular}{rr|r}
$A$ & $B$ & $A \times B$ \\
\hline
\hline
-1  & -1  & 1            \\
-1  & 1   & -1           \\
1   & -1  & -1           \\
1   & 1   & 1           
\end{tabular}
\quad\quad\quad\quad
\quad\quad\quad\quad
\begin{tabular}{rr|r}
$A$ & $B$ & $\overline{A \oplus B}$ \\
\hline
\hline
0  & 0  & 1            \\
0  & 1   & 0           \\
1   & 0  & 0           \\
1   & 1   & 1           
\end{tabular}
\end{table}

The binary and ternary $\tanh$ activation functions are implemented by testing the sign (in the case of binary $\tanh$) or sign and magnitude (for ternary $\tanh$) of the input and yielding the corresponding value $\pm 1$ or $0$ as seen in Fig.~\ref{fig:activation_binary_ternary}.
A binary or ternary $\tanh$ activation layer preceded by a batch normalization (BN) layer~\cite{Ioffe:2015:BNA:3045118.3045167} can be further optimized.
The BN layer shifts the output of the dense layers to the range of values in which the activation function is nonlinear, enhancing the network's capability of modeling nonlinear responses. The usual BN transformation $y$ for an input $x$ is

\begin{equation}
    y = \frac{x - \mu}{\sqrt{\sigma ^ 2 + \epsilon}}\gamma + \beta,
\end{equation}

given the mean $\mu$, variance $\sigma^2$, scale $\gamma$, and shift $\beta$ learned during the network training.
For a  BN followed by a binary $\tanh$ activation, the sign of $y$ is enough to determine a node output value. To avoid calculating the scaling of $x$ using FPGA DSPs, the four BN parameters are used to compute the value of $x$ at which $y$ flips sign. 
This calculation is performed at compilation time, when the model is converted to HLS firmware using \texttt{hls4ml}.
Similarly, the two values of $x$ around which the output of the ternary $\tanh$ activation changes are also calculated at compilation time.
In the FPGA, each node output is then simply compared against these precomputed thresholds, outputting the corresponding $\pm 1$, or $0$.
An additional optimization step sets the type of $x$ in the HLS implementation to integer with a bit width corresponding to the largest integer expected for each binary/ternary layer, found at compilation time. This procedure further saves FPGA resources.

The binary and ternary layers considered for this work are fully integrated and compatible with the {\tt hls4ml} package. While not explored here, the package also supports models mixing binary/ternary layers with higher precision layers for fully customized networks.

\section{Binarization and ternarization strategies}
\label{sec:architectures}

Given a full-precision model, one could follow different strategies to turn it into a binary or ternary model. One could just replace each full-precision component by the corresponding binary/ternary element, in order to minimize resource utilization. This might result in a loss of accuracy. As an alternative, one could train a binary/ternary model with arbitrarily large architecture, in order to match the accuracy obtained at full precision, at a cost of a larger latency and resource consumption. The ultimate strategy to follow depends on the use case. In this work, we present a few options, covering these two extremes and intermediate solutions. 

\begin{figure}[t!]
    \centering
    \includegraphics[width=0.8\textwidth]{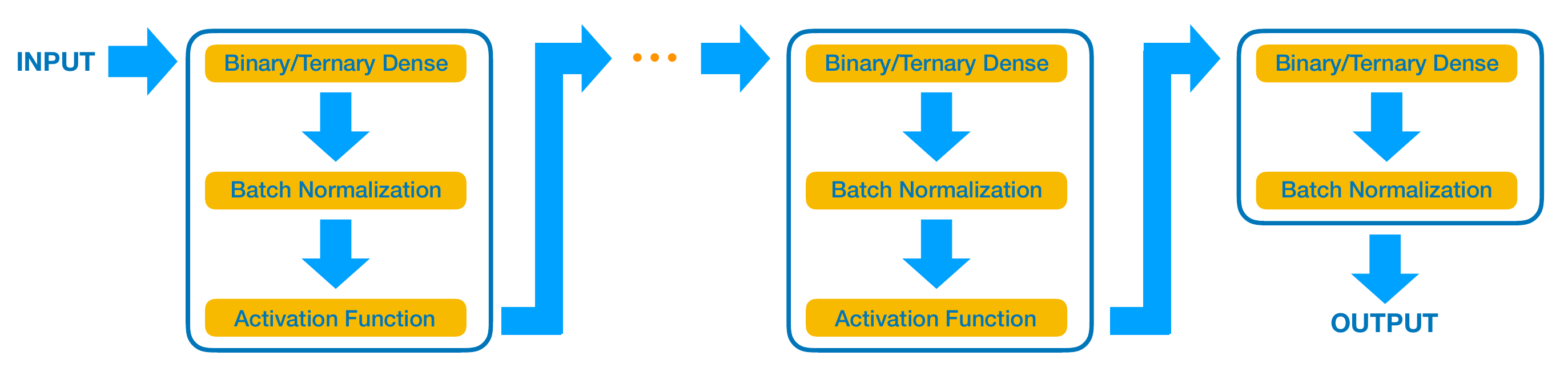}
    \caption{The MLP architecture used in this study, consisting of a sequence of repeating blocks. Each block, fully connected to the previous and following one, consists of a dense layer, a batch normalization layer, and an activation layer. The last block does not have an activation layer.\label{fig:architecture}}
\end{figure}

In this work, we focus on binary/ternary MLPs. The basic structure for the adopted architectures is shown in Fig.~\ref{fig:architecture}. Each model consists of a sequence of blocks, each composed of a dense, BN, and activation layer.
For binary and ternary $\tanh$, a BN+activation layer sequence can be implemented at small resource cost (see Section~\ref{sec:FPGAopt}), which makes this choice particularly convenient for fast inference on edge devices. 

The binarization/ternarization of a given model can be done in different ways, e.g., preserving the model architectures or its performance. As a consequence, for each benchmark problem we consider seven models:
\begin{itemize}
  \item {\it Baseline}: the three-layer MLP described in Section~\ref{sec:data}.

  \item {\it Binarized (BNN)}: a binary version of the baseline model,
    built preserving the model architecture (number of layers and
    nodes) while applying the following changes: use a binary
    representation ($\pm 1$) for the weights; replace the
    inner-layer ReLU activation functions with a binary $\tanh$
    (see Fig.~\ref{fig:activation_binary_ternary}); introduce BN layers in between the binary dense layers and the
    activation functions; remove the softmax activation function in
    the output layer.

  \item {\it Ternarized (TNN)}: a ternary version of the baseline model,
    built preserving the model architecture (number of layers and
    nodes) while applying the following changes: use a ternary
    representation ($-1,0,+1$) for the weights; replace the inner-layer
    ReLU activation functions with a ternary $\tanh$ (see
    Fig.~\ref{fig:activation_binary_ternary}); introduce BN layers in between the ternary dense layers and the
    activation functions; remove the softmax activation function in
    the output layer.

  \item {\it Best BNN}: same structure as the BNN model,
    but with more nodes in each layer to improve performance. We obtain this model with a Bayesian optimization performed using {\tt GPyOpt}~\cite{gpyopt2016}, finalized to minimize the validation loss in the training process.

   \item {\it Best TNN}: same structure as the TNN
     model, but with the number of nodes per layer chosen through a
     Bayesian optimization of the architecture, as for the best BNN
     model. 

   \item {\it Hybrid BNN}: same as the BNN model, but with
     ReLU or clipped ReLU activation functions rather than the binary $\tanh$ of
     Fig.~\ref{fig:activation_binary_ternary}.

   \item {\it Hybrid TNN}: same as the TNN model, but with ReLU or clipped ReLU activation functions rather than the ternary $\tanh$ of
     Fig.~\ref{fig:activation_binary_ternary}.     
\end{itemize}    

The baseline model is taken as a benchmark of ideal performance and the other models represent different strategies toward a more resource-friendly representation. The BNN and TNN models are simple translations of the baseline model. They are designed to reduce the resource consumption, at the potential cost of a performance drop. The best models are designed to match (as close as possible) the performance of the baseline model, which might result in a larger resource consumption with respect to what the BNN and TNN models achieve. The hybrid models are a compromise between the two approaches. The fixed-precision conversion is applied only to the weights and biases of the nodes in the dense layers, while ReLU or clipped ReLU activation functions are used. Given the relatively small resources used by the ReLU/clipped ReLU activations, the hybrid models allow one to reach performance closer to the baseline model without inflating the number of nodes and, consequently, numerical operations. The best BNN and TNN models are only presented for the LHC jet problem, since in that case the simple binarization or ternarization of the baseline model result in a substantial performance loss. The effect is much milder for the MNIST classification problem, so that the binary and ternary architectures are not re-optimized for in that case.

Not all of the operations or intermediate outputs of a binary (ternary) are represented in binary (ternary) precision, e.g., the output of a ReLU activation function in a hybrid model. For this reason, in the following we discuss bit precision and network quantization even in the context of binary and ternary models.

All models are implemented in {\tt Keras}~\cite{keras}, with {\tt TensorFlow}~\cite{TF} as a backend using the implementation in~\cite{bertmoons} for binary and ternary layers, which we also cross-checked with {\tt QKeras}~\cite{qkeras} with similar results. The network training was performed on an NVIDIA Tesla V100 GPU. During training, binary/ternary precision is employed during forward propagation, while full precision is used during backward propagation. The baseline models of Section~\ref{sec:data} are trained minimizing a categorical cross entropy. The binary and ternary models are trained minimizing a hinge loss function~\cite{Lin:2002}. While the hinge loss has been found to give the best performance for binary/ternary networks~\cite{courbariaux,binary_first,ternary_first}, the same choice for the baseline models is arbitrary. We have verified that the baseline models trained with the hinge loss after replacing the last softmax layer (Fig.~\ref{fig:input_arch}) with a dense plus batch normalization layers yield similar results in terms of both accuracy and resource usage.

\section{Experiments}
\label{sec:exp}

The results presented below are synthesized with the Vivado HLS version 2018.2 for a Xilinx Virtex Ultrascale 9+ FPGA with part number xcvu9p-flga2104-2L-e. The clock frequency is fixed at 200 MHz, which is typical for the LHC L1 triggers. For this configuration we study the FPGA resources used by the models described in Section~\ref{sec:architectures}. There are four main resource categories: the on-board FPGA memory (BRAM), DSPs, and registers and programmable logic (flip-flops, or FFs, and lookup tables, or LUTs). Unless otherwise specified, the quoted results are derived after the HLS compilation step. The network implementation is further refined by the logic synthesis. This step transforms the Register Transfer Level (RTL) design created by the HLS compiler into a gate-level implementation, applying additional optimizations that result in a more accurate assessment of the resource utilization. We verified that this final step does not affect the accuracy while it reduces the resource consumption. 

All results quoted in this section are taken from the numerical simulation of the synthesized firmware. This numerical simulation is one of the tools provided by the FPGA vendor and gives bit-identical results to running on a physical device. On the other hand, running on a physical device is a much more consuming operation. Given the large number of tests considered in this study, we omitted this last step, mainly for practical reasons.

\subsection{Handwritten digits classification}

We first evaluate the performance of the HLS neural network implementation for the models described in Section~\ref{sec:architectures} with different fixed-point precisions by scanning the number of both integer (I) and fractional (F) bits. In the following, a given choice of fixed-point precision is specified as $\langle \mathrm{T}, \mathrm{I} \rangle$, where $\mathrm{T}=\mathrm{I}+\mathrm{F}$ is the total number of allocated bits.
For each case, the minimum number of bits yielding an accuracy above 90\% after quantization is considered.
We then study the latency and resource utilization in these configurations. 
Table~\ref{tab:quantizationMNIST} shows a comparison of the performance obtained for the baseline, binary, and ternary models, in terms of accuracy and AUCs, before and after quantization.

\begin{table}[ht!]
\centering
\caption{Accuracy and AUCs of the different MNIST-classification models described in Section~\ref{sec:architectures} before and after quantization, for the fixed point precision settings chosen for this study. Both the numbers of integer (I) and fractional (F) bits are specified, using the notation $\langle \mathrm{I}+\mathrm{F}, \mathrm{I} \rangle$. For each case, the AUCs are reported as the range spanned by the classes with lowest and highest identification performance.\label{tab:quantizationMNIST}}
\begin{tabular}{c|cc|ccc}
\hline
Model & \multicolumn{2}{c|}{Floating point precision} & \multicolumn{3}{c}{Fixed point precision}\\
& AUC & Accuracy [\%] & Number of bits & AUC & Accuracy [\%] \\
\hline
\hline
Baseline & 0.9991--0.9997 & 98 & $\langle 18, 8 \rangle$ & 0.9919--0.9959 & 95\\
BNN      & 0.9869--0.9979 & 93 & $\langle 16, 8 \rangle$ & 0.9860--0.9976 & 93\\
TNN      & 0.9921--0.9992 & 95 & $\langle 16, 6 \rangle$ & 0.9918--0.9992 & 95\\
\hline
Hybrid BNN (ReLU)    & 0.9953--0.9990 & 95 & $\langle 16, 10 \rangle$ & 0.9956--0.9989 & 95\\
Hybrid TNN (ReLU)    & 0.9970--0.9993 & 96 & $\langle 16, 10 \rangle$ & 0.9971--0.9993 & 96\\
Hybrid BNN (clipped ReLU) & 0.9827--0.9983 & 95 & $\langle 16, 10 \rangle$ & 0.9828--0.9983 & 95\\
Hybrid TNN (clipped ReLU) & 0.9857--0.9989 & 96 & $\langle 16, 10 \rangle$ & 0.9859--0.9988 & 96\\
\hline
\end{tabular}
\end{table}

\begin{figure}[t!]
\centering
\includegraphics[width=0.48\textwidth]{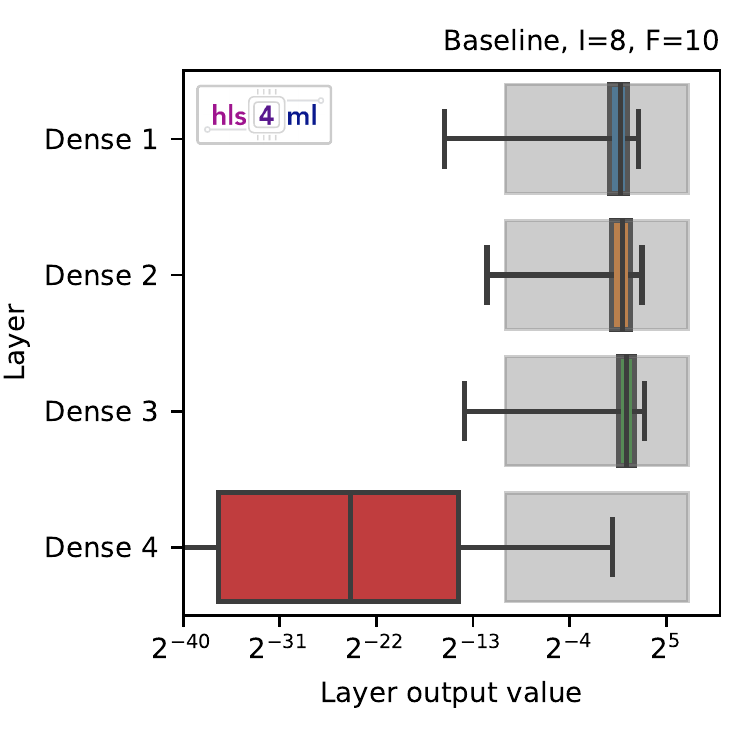}
\includegraphics[width=0.48\textwidth]{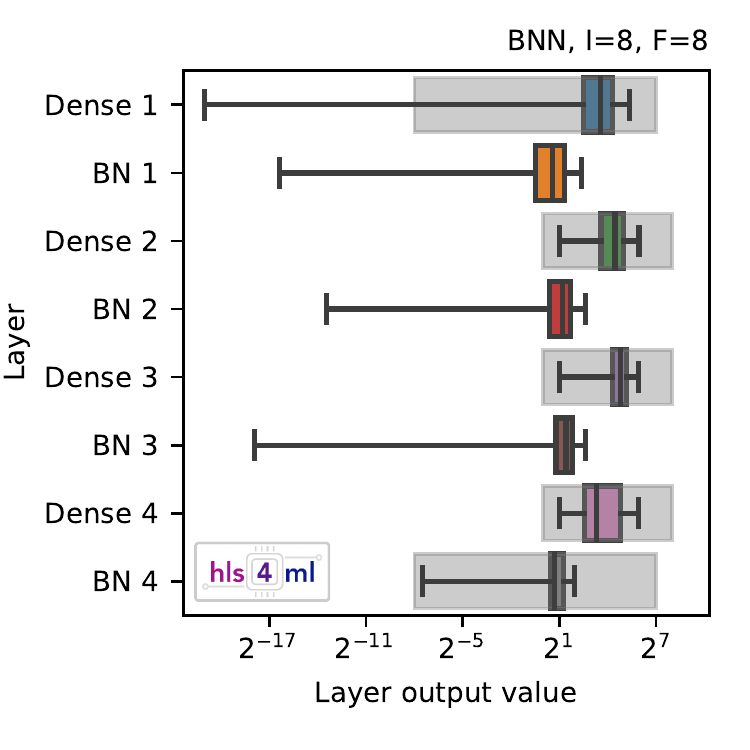}
\caption{Profile of the range of output values of each layer, sampled during inference on the test data set, for the baseline (left) and BNN (right) MNIST models. For each layer, the box represents the quartiles of the distribution, while the line shows the median.
The lines extending beyond the box show the minimum and maximum values. The gray shaded areas represent the range covered by the allocated fixed point precision for each layer. In the left plot, these ranges  correspond to the precision specified at compilation ($\langle 18, 8 \rangle$). On the right plot, an optimization procedure implemented in {\tt hls4ml} for binary and ternary networks automatically adapts the precision of each layer to match the range covered by the output distribution; as the BN layer is merged with the binary $\tanh$ in the HLS implementation, its output precision is 1 bit. Dense, batch normalization (BN), and activation layers are presented in order from the input (top) to the output (bottom). \label{fig:NNprofiles_0}}
\end{figure}

For binary and ternary models, the {\tt hls4ml} library applies a further level of per-layer customization of the fixed-point representation, to match the numerical precision of each layer separately, as discussed in Section~\ref{sec:FPGAopt}. The outcome of this optimization is shown in the right plot of Fig.~\ref{fig:NNprofiles_0} for the BNN model, where the gray areas cover different numerical ranges for different layers, despite the common precision specified at compilation ($\langle 16, 8 \rangle$ in this case). During the optimization, the inputs and the outputs are still represented by the fixed-point precision specified by the user, while the precision of the other network components is optimized.

When quantizing a model, one should allocate I and F bits so that the range of values one can cover overlaps with the range of values returned by the network layers, in order to reduce the impact on accuracy. This is shown in the left plot of Fig.~\ref{fig:NNprofiles_0}, where the profile of output values returned by each layer of the baseline model is compared to the range covered by the allocated fixed-point precision. For each layer, we consider the distribution of the output values obtained running the network on a test sample. In the figure, the box represents the quartiles of the distribution, while the line inside the box shows the median. The lines extending beyond the box show the minimum and maximum values. The gray area represents the numerical range covered by the allocated precision. Overall, the optimized precision matched the bulk of the output values at each layer. The only exception is observed for the output layer. In this case, the allocated precision (gray area in the last row of the left plot in Fig.~\ref{fig:NNprofiles_0}) does not cover the bulk of values returned by the layer (red box in the figure). This happens whenever a given example is associated to a specific class with a score close to 1, so that the other values  are pushed close to 0 and out of the supported range. In practice, this fact would not alter the classification outcome in inference. For instance, this would not be a problematic aspect when operating this algorithm through the $\argmax$ function, associating a given example to the class with the largest output.

For the baseline model, the quantization from floating-point precision to $\langle 18, 8 \rangle$ results in an accuracy drop from 98\% to 95\%. 
This is almost entirely induced by the softmax activation function applied to the last layer and it results from the limited precision of the LUT implementing the $\exp$ functions in the softmax. This parameter is hard-coded in the version of {\tt hls4ml} used for this study. One could avoid this accuracy loss by removing the softmax function at the end of the HLS implementation of the inference, as long as there is interest only on which class has the biggest score and not on the individual scores. An alternative option is to further optimize the precision of the LUT implementing the softmax activation function. In this case, we verified that a $\langle 18, 8 \rangle$ quantization baseline with $\langle 22, 10 \rangle$ precision for the softmax LUT recovers an accuracy of 97\% without affecting the resources. The ability to externally configure the precision of the softmax LUT will be implemented in future versions of {\tt hls4ml}.

\begin{figure}[t!]
\centering
\includegraphics[width=0.48\textwidth]{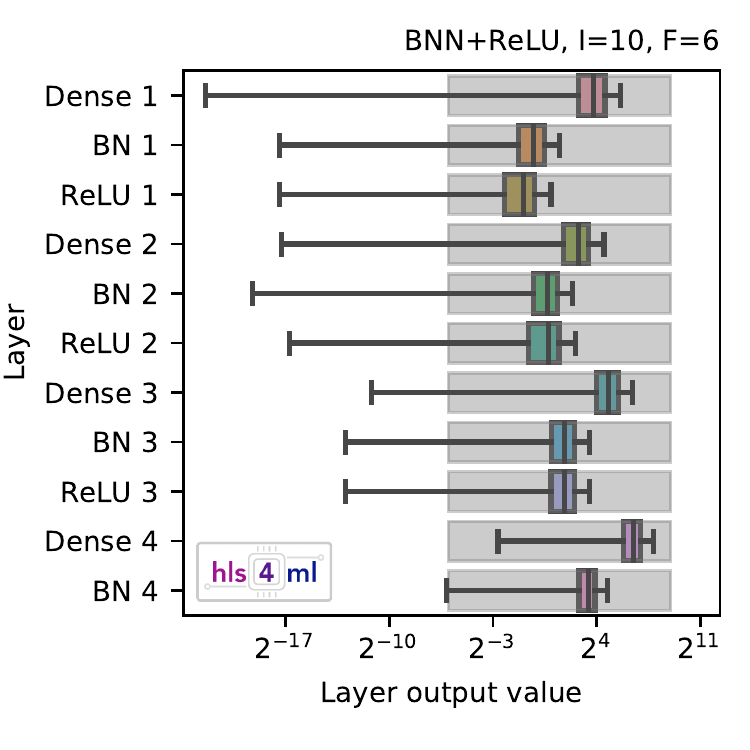}
\includegraphics[width=0.48\textwidth]{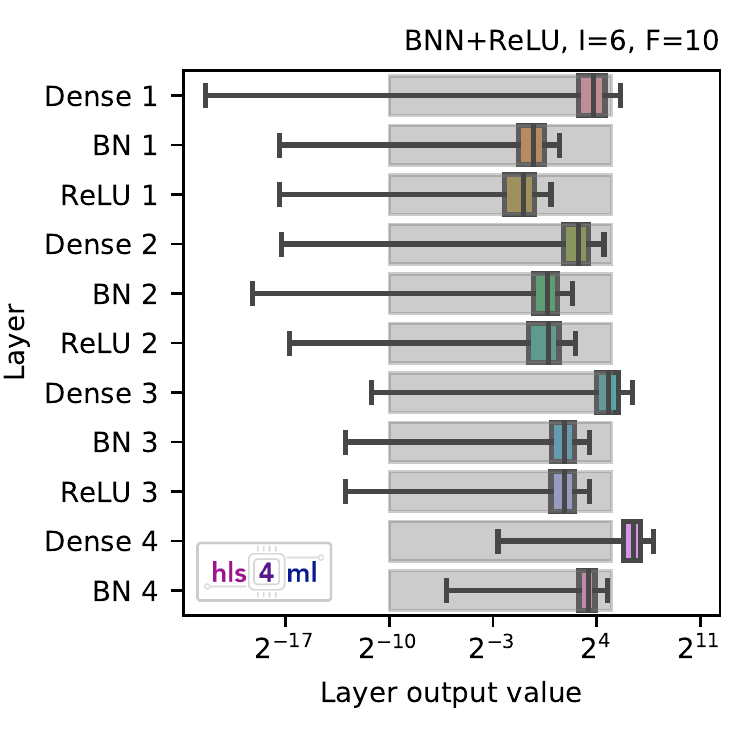}
\caption{Profile of the range of output values of each layer, sampled during inference on the test data set, for the hybrid BNN+ReLU model quantized to 16-bit precision, when 10 (left) or 6 (right) bits are used for the integer part. 
For each layer, the box represents the quartiles of the distribution, while the line shows the median.
The lines extending beyond the box show the minimum and maximum values.
The gray shaded areas represent the range covered by the allocated fixed-point precision for each layer. Dense, batch normalization (BN), and activation layers are presented in order from the input (top) to the output (bottom).\label{fig:NNprofiles_1}}
\end{figure}

For the hybrid BNN/TNN models, the same number of bits used for the BNN/TNN cases allows one to achieve the FPP accuracy, at the condition of allocating more integer (10 instead of 6) and less fractional (6 instead of 10) bits. This behaviour can be understood from Figure~\ref{fig:NNprofiles_1}, which shows the range of outputs returned by each hybrid BNN layer. While for I=10 the allocated precision spans the full range of outputs returned by each layer, frequent overflows are observed for the Dense 1, Dense 3 and Dense 4 layers when we set I=6. 


\begin{table}[ht!]
\centering
\caption{Comparison of the resource utilization for the MNIST-classification models described in Section~\ref{sec:architectures}, together with timing information. Resources estimated by the HLS compiler (C) and obtained by the logic synthesis (S) are quoted for a chosen initiation interval (II).\label{tab:resourcesMNIST}}
\begin{tabular}{c|cccccccccc}
\hline
Model & II & Latency [ns] & \multicolumn{2}{c}{DSPs [\%]} &
\multicolumn{2}{c}{FFs [\%]} & 
\multicolumn{2}{c}{LUTs [\%]} & 
\multicolumn{2}{c}{BRAMs [\%]} \\
 & & & 
 C & S &
  C & S &
  C & S &
 C & S \\
\hline
\hline
Baseline & 28 & 315 & 130 & 100 & 18 & 8 & 69 & 54 & 126 & 61 \\
BNN  & 14 & 200 & 0 & 0 & 5 & 7 & 155 & 18 & 46 & 16 \\
TNN  & 14 & 190 & 0 & 0 & 6 & 7 & 174 & 22 & 52 & 16 \\
\hline
Hybrid BNN (ReLU)         & 14 & 200 & 1 & 0.16 & 7 & 9  & 215 & 31 & 52 & 16 \\
Hybrid TNN (ReLU)         & 14 & 200 & 1 & 1 & 7 & 10 & 217 & 35 & 52 & 16 \\
Hybrid BNN (clipped ReLU) & 14 & 200 & 1 & 2 & 7 & 8  & 215 & 29 & 52 & 16 \\
Hybrid TNN (clipped ReLU) & 14 & 200 & 1 & 1 & 7 & 9  & 215 & 31 & 52 & 16 \\
\hline
\end{tabular}
\end{table}

Table~\ref{tab:resourcesMNIST} provides a comparison of the resource utilization and latency for the configurations presented in Tab.~\ref{tab:quantizationMNIST}. For each configuration, we quote both the resource utilization estimated by the HLS compiler and those obtained by the logic synthesis. In the table, the II represents the number of clock cycles needed before the algorithm may accept a new set of inputs. In our study, the II value is fixed by requiring that the resulting resource utilization is below the maximum allowed on the target FPGA.  Lower II values would result in a network design that would not fit the device. Larger II values would result in higher latency.

At the very low latency values (${\cal O}(100)$~ns) that we are targeting, BNN/TNN models allow one to reach competitive performance while saving most of the FPGA resources. About half of the observed accuracy loss can be recovered using hybrid BNN/TNN models, paying a small price in terms of DSPs utilization, induced by an explicit allocation of a BN layers before the ReLU/clipped ReLU activation functions rather than the bit-shift implementation described in Section~\ref{sec:FPGAopt}. A further optimization of the BN operations for hybrid models could in principle 
push the DSPs utilization closer to zero.

The LUTs usage is largely overestimated by the HLS compiler for all binary and ternary NN models, while it is found to be well within the available resources after the logic synthesis. Hybrid models require more LUTs with respect to the standard BNN/TNN, because of the wider data bit width at the input of each binary or ternary layer.

\begin{figure}[t!]
    \centering
    \includegraphics[width=0.45\textwidth]{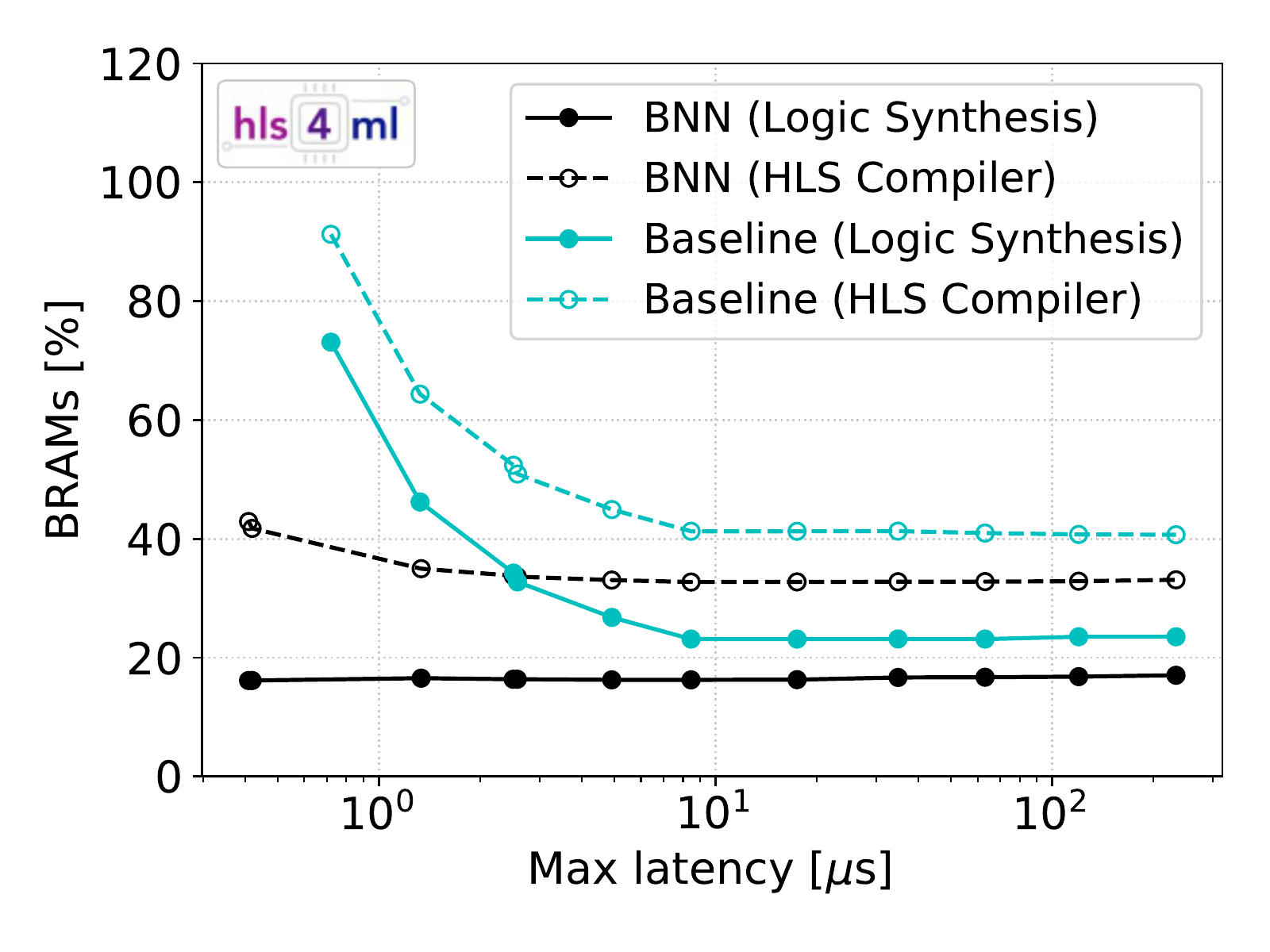}
    \includegraphics[width=0.45\textwidth]{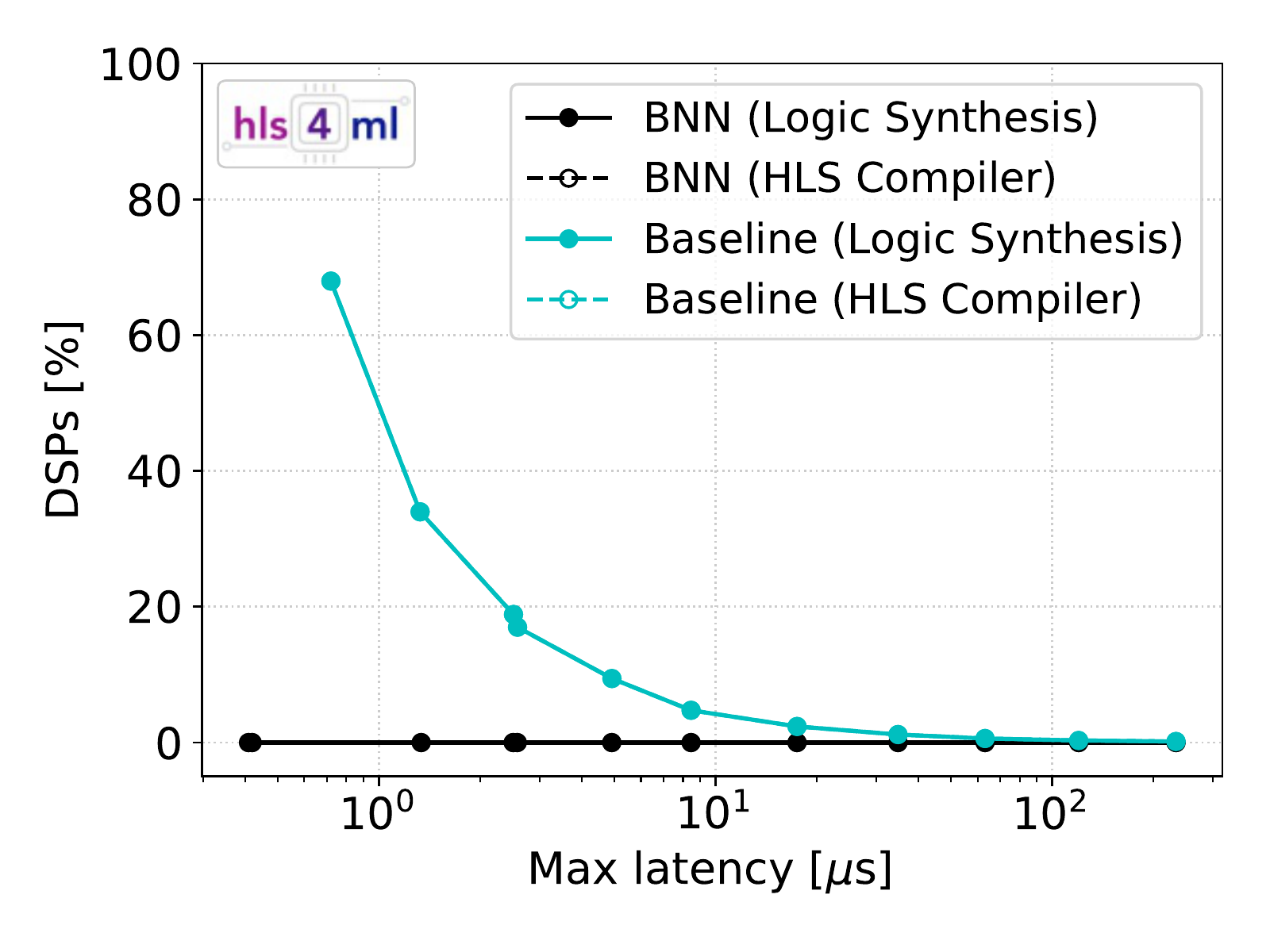}\\
    \includegraphics[width=0.45\textwidth]{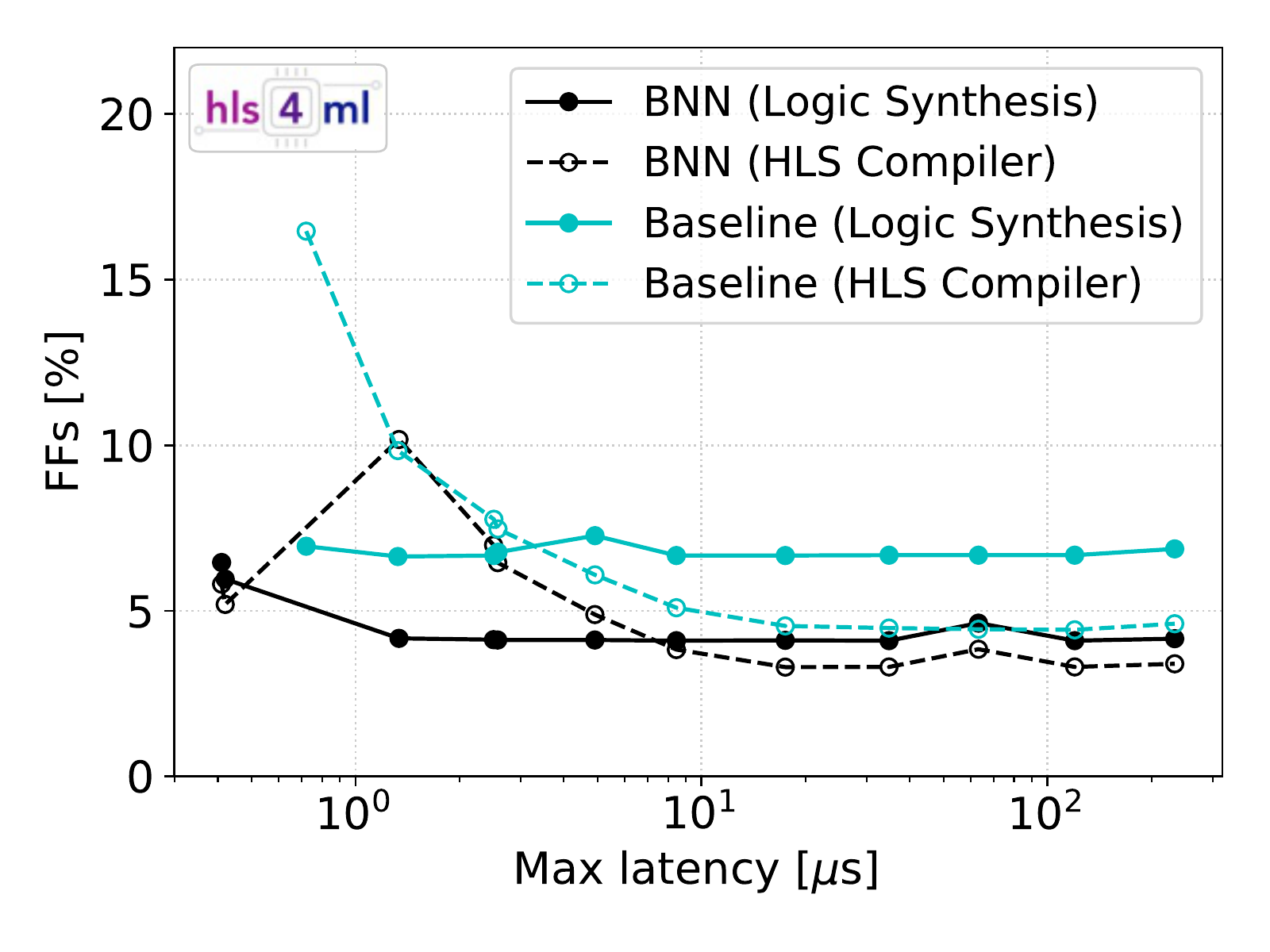}
    \includegraphics[width=0.45\textwidth]{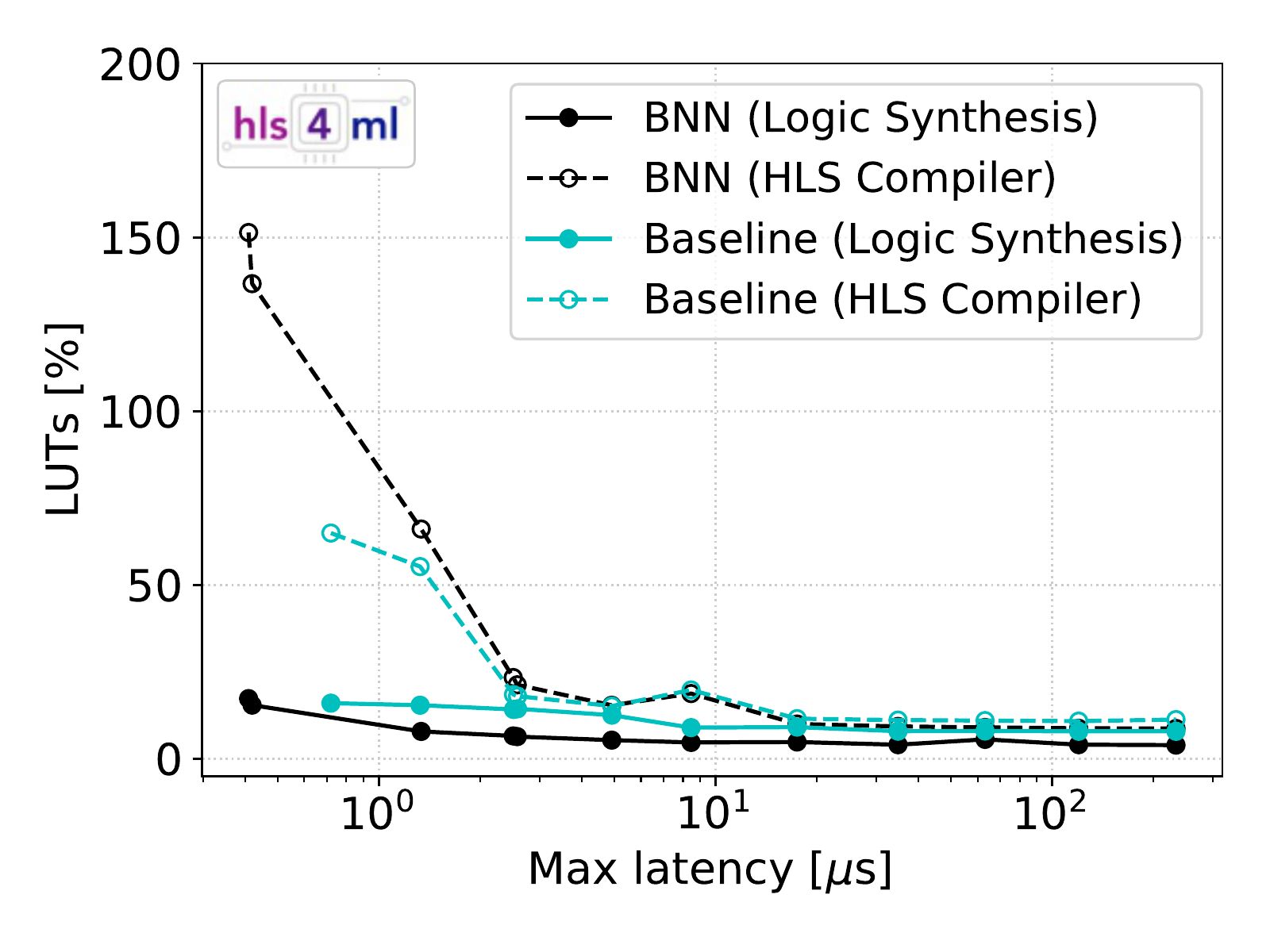}
    \caption{Comparison of the resource utilization estimated by the HLS compiler and obtained by the logic synthesis versus the maximum latency achieved by the design for the BNN and baseline MNIST-classification models. The TNN model gives similar resource utilization as the BNN and is omitted.\label{fig:scans}}
\end{figure}

Figure~\ref{fig:scans} shows the dependence of the resource utilization on the maximum latency achieved by the design (controlled by the II) for the baseline and BNN models. Results for the TNN model are very close to the BNN ones. For all latency values, the resources used by the BNN/TNN models are typically reduced with respect to the baseline model. In particular, the number of DSPs used is greatly reduced for latency values up to a few $\mu$s. For higher latency values, the II is large enough to allow a small usage of DSPs even for the baseline model. In that case, the advantage of using a binary or ternary quantization would be minor. Due to technical aspects of the implementation of very-wide dense layers in {\tt hls4ml}, it is not possible to configure the model to run with smaller latency values than those shown.

As a final test, we train a larger BNN model consisting of three dense layers with 256 nodes each, as in the study of Ref.~\cite{xilinx_finn}, allowing for a direct comparison of our implementation of a binary architecture with what presented there. The {\tt hls4ml} implementation of this model yields a total accuracy of 95\% for both floating-point and fixed-point precision, where the latter is fixed to $\langle 16, 6 \rangle$. With an II of 28, we obtain a maximum latency of 0.31~$\mu$s with a resource utilization comparable to that in Ref.~\cite{xilinx_finn}. In particular, the deployed model obtained with {\tt hls4ml} after the  logic synthesis utilizes 0\% DSPs, 7\% FFs, 23\% LUTs, and 16\% BRAMs on a Xilinx Virtex Ultrascale 9+ FPGA card.

\subsection{LHC jet identification}
\label{sec:modelsLHC}

As a second benchmark example, we consider the LHC jet-tagging problem introduced in Section~\ref{sec:data} and study all the binarization/ternarization strategies described in Section~\ref{sec:architectures}. 
For all models a fixed-point precision of $\langle 16, 6 \rangle$ is sufficient to reproduce the FPP accuracy after quantization.
The AUCs and accuracy before and after quantization are reported in Table~\ref{tab:quantizationJets} for all models, while a comparison of the resource utilization is found in Table~\ref{tab:resourcesJets}.

\begin{table}[htb]
\centering
\caption{Accuracy and AUCs of the different LHC jet tagging models described in Section~\ref{sec:architectures} before and after quantization, for fixed-point precision $\langle \mathrm{I}+\mathrm{F}, \mathrm{I} \rangle$ chosen for this study. For each case, the AUCs are reported as the range spanned by the classes with lowest and highest identification performance.\label{tab:quantizationJets}}
\footnotesize
\begin{tabular}{c|c|cc|ccc}
\hline
Model & Architecture & \multicolumn{2}{c|}{Floating point precision} & \multicolumn{3}{c}{Fixed point precision}\\
      &              & AUC & Accuracy [\%]                           & Number of bits & AUC & Accuracy [\%]\\
\hline
\hline
Baseline & $16{\times}64{\times}32{\times}32{\times}5$ & 0.904--0.958 & 75 & $\langle 16, 6 \rangle$ & 0.900--0.955 & 75\\
BNN      & $16{\times}64{\times}32{\times}32{\times}5$ & 0.794--0.891 & 58 & $\langle 16, 6 \rangle$ & 0.794--0.891 & 58\\
TNN      & $16{\times}64{\times}32{\times}32{\times}5$ & 0.854--0.915 & 67 & $\langle 16, 6 \rangle$ & 0.854--0.915 & 67\\
\hline
Best BNN & $16{\times}448{\times}224{\times}224{\times}5$  & 0.886--0.937 & 72 & $\langle 16, 6 \rangle$ & 0.884--0.938 & 72\\
Best TNN & $16{\times}128{\times}64{\times}64{\times}64{\times}5$ & 0.886--0.931 & 72 & $\langle 16, 6 \rangle$ & 0.886--0.930 & 72\\
\hline
Hybrid  BNN (ReLU) & $16{\times}64{\times}32{\times}32{\times}5$ & 0.862--0.920 & 69 & $\langle 16, 6 \rangle$ & 0.862--0.919 & 69\\
Hybrid  TNN (ReLU) & $16{\times}64{\times}32{\times}32{\times}5$ & 0.874--0.934 & 70 & $\langle 16, 6 \rangle$ & 0.874--0.934 & 70\\
\hline
Hybrid  BNN & \multirow{2}{*}{$16{\times}64{\times}32{\times}32{\times}5$} & \multirow{2}{*}{0.852--0.916} & \multirow{2}{*}{67} & \multirow{2}{*}{$\langle 16, 6 \rangle$} & \multirow{2}{*}{0.852--0.916} & \multirow{2}{*}{67} \\
(clipped ReLU) & & & & & & \\
Hybrid  TNN  & \multirow{2}{*}{$16{\times}64{\times}32{\times}32{\times}5$} & \multirow{2}{*}{0.874--0.921} & \multirow{2}{*}{70} & \multirow{2}{*}{$\langle 16, 6 \rangle$} & \multirow{2}{*}{0.874--0.921} & \multirow{2}{*}{70}\\
(clipped ReLU) & & & & & & \\

\hline
\end{tabular}
\end{table}

Unlike what is seen for the MNIST digit classification, the simple binarization/ternarization of the baseline model results in a big accuracy loss. This is partially mitigated by the use of ReLU and clipped ReLU activations. As an alternative approach, we also consider 
optimized binary and ternary architectures (best models in Table~\ref{tab:quantizationJets}), fixed through a Bayesian optimization of the network hyperparameters. The result of the Bayesian hyperparameter optimization for BNN and TNN converges to architectures with about 40 and 4 times more parameters with respect to the baseline architecture, respectively. With these larger architectures, binary and ternary methods almost match, with a moderate loss in accuracy. Optimizing the architecture of the binary and ternary models yields comparable precisions, but with a different resource balance (e.g., DSPs vs. LUTs), offering an alternative that might better fit certain use cases. 

The results of Tables~\ref{tab:quantizationJets}~and~\ref{tab:resourcesJets} confirm that ternary networks generally offer a better resource vs. accuracy balance than binary networks, with a minimal (often negligible) additional resource cost and a comparable (sometimes smaller) latency. In terms of FPGA resources, even the large architecture of the best TNN model results in a limited resource usage, well below the baseline model. Instead, the largest best BNN model requires a higher II value to fit the FPGA resource boundaries. The latency is kept within the $\sim 1 \mu$s boundary we target, but is significantly larger than what is achieved by the best TNN and the baseline models. The best TNN model gives the same accuracy as the best BNN model, with the same latency as the baseline model but with a drastic reduction of DSP utilization .

\begin{table}[htb]
\centering
\caption{Comparison of the resource utilization for the LHC jet-tagging models described in Section~\ref{sec:architectures}, together with timing information. Resources estimated by the HLS compiler (C) and obtained by the logic synthesis (S) are quoted for a chosen initiation interval (II).\label{tab:resourcesJets}}
\begin{tabular}{c|cccccccccc}
\hline
Model & II & Latency [ns] & \multicolumn{2}{c}{DSPs [\%]} &
\multicolumn{2}{c}{FFs [\%]} & 
\multicolumn{2}{c}{LUTs [\%]} & 
\multicolumn{2}{c}{BRAMs [\%]} \\
 & & & 
 C & S &
  C & S &
  C & S &
 C & S \\
\hline
\hline
Baseline & 1 & 60 & 60 & 57 & 1 & 1 & 7 & 5 & 0 & 0\\
BNN      & 1 & 40 & 0  & 0  & 0 & 0 & 3 & 1 & 0 & 0\\
TNN      & 1 & 40 & 0  & 0  & 0 & 0 & 4 & 1 & 0 & 0\\
\hline
Best BNN & 16 & 205 & 0 & 0 & 1 & 3 & 128 & 8 & 12 & 0\\
Best TNN & 1  & 55  & 0 & 0 & 0 & 0 & 14  & 3 & 0  & 0\\
\hline
Hybrid  BNN (ReLU)    & 1 & 50 & 2 & 2 & 0 & 0 & 6 & 2 & 0 & 0\\
Hybrid  TNN (ReLU)    & 1 & 50 & 2 & 2 & 0 & 0 & 7 & 2 & 0 & 0\\
Hybrid  BNN (clipped ReLU) & 1 & 50 & 2 & 2 & 0 & 0 & 6 & 2 & 0 & 0\\
Hybrid  TNN (clipped ReLU) & 1 & 50 & 2 & 2 & 0 & 0 & 7 & 2 & 0 & 0\\
\hline
\end{tabular}
\end{table}

\section{Summary and Outlook}
\label{sec:conclusions}

We presented the implementation of binary and ternary networks in the {\tt hls4ml} library, designed to automatically convert a given neural network model into firmware of an FPGA card. Using two benchmark classification examples (handwritten digit recognition on the MNIST data set and jet identification at the LHC), we discuss different strategies to convert a given model into a binary or a ternary model. We showed how binary and ternary networks allow one to preserve competitive performance (in terms of accuracy) while drastically reducing the resource utilization on the card and, at the same time, keeping the inference latency at ${\cal O}(100)$~ns. When compared to binary models, ternary models reach accuracy values much closer to the original baseline models, at a typically smaller resource cost and comparable latency. Model binarization and ternarization are competitive alternatives to other compression approaches (e.g., pruning) and represent the ultimate resource saving in terms of network quantization. They offer a qualitative advantage of keeping DSP utilization at a minimum, and offer an interesting opportunity to deploy complex architectures on resource constrained environments, such as the L1 trigger system of a typical collider physics experiment.

\section*{Acknowledgement}

 We acknowledge the Fast Machine Learning collective  (\url{https://fastmachinelearning.org}) as an open community of multi-domain experts and collaborators. This community was important for the development of this project. 

M.~P., S.~S., V.~L. and J.~N. are supported by the European Research Council (ERC) under the European Union's Horizon 2020 research and innovation program (grant agreement n$^o$ 772369).
S.~J., M.~L., K~.P., and N.~T. are supported by Fermi Research Alliance, LLC under Contract No. DE-AC02-07CH11359 with the U.S. Department of Energy, Office of Science, Office of High Energy Physics.
P.~H. is supported by a Massachusetts Institute of Technology University grant. PH and DR thank support from NSF AWARD \#190444, \#1934700, \#1931469, \#1836650. Z.~W. is supported by the National Science Foundation under Grants No. 1606321 and 115164.

\section*{Data availability}

The data that support the findings of this study are openly available at \url{https://doi.org/10.5281/zenodo.3602260} (LHC jet dataset) and \url{https://www.openml.org/d/554} (MNIST handwritten digit database).

\bibliographystyle{lucas_unsrt}  
\bibliography{references}

\end{document}